# Engineering Reasoning and Instruction (ERI) Benchmark: A Large Taxonomy-driven Dataset for Foundation Models and Agents

M.Z. Naser[1,2], Ahmad Bani Awwad[1], Zoie McCreery[1], Radwa Eissa[1], Ahmad Naser[3], , Gianluca Cusatis[4], Andrew Metcalf[5], Kapil Madathil[6], Jamal Abdalla[7], Venkatesh Kodur[8], Mohammad Reza Saeb[9]

[1]School of Civil and Environmental Engineering & Earth Sciences, Clemson University, USA
E-mail: mznaser@clemson.edu, abaniaw@clemson.edu, zmccree@g.clemson.edu, reissa@clemson.edu
[2]Artificial Intelligence Research Institute for Science and Engineering, Clemson University, USA
[3]Department of Mechanical Engineering, University of Manitoba, Canada
E-mail: a.naser@umanitoba.ca
[4]Department of Civil and Environmental Engineering, Northwestern University, USA
Email: g-cusatis@northwestern.edu
[5]Department of Environmental Engineering and Earth Sciences, Clemson University, USA
E-mail: ametcal@clemson.edu
[6]Departments of Industrial Engineering, Clemson University, USA
E-mail: kmadath@clemson.edu
[7]Departments of Civil Engineering, American University of Sharjah, UAE
E-mail: jabdalla@aus.edu
[8]Departments of Civil and Environmental Engineering, Michigan State University, USA
Email: kodur@egr.msu.edu
[9]Department of Pharmaceutical Chemistry, Medical University of Gdańsk, Gdańsk, Poland
Email: mr.saeb@gumed.edu.pl

## Abstract

The Engineering Reasoning and Instruction (ERI) benchmark is a taxonomy-driven instruction dataset designed to train and evaluate engineering-capable large language models (LLMs) and agents. This dataset spans nine engineering fields (namely: civil, mechanical, electrical, chemical, environmental, aerospace, materials, fire, and industrial engineering) and 55 subdomains, and is crossed with seven intent types (i.e., definition, explanation, calculation, comparison, design/synthesis, troubleshooting, and code-related) and three difficulty tiers (undergraduate, graduate, and professional), yielding 57,750 records with field/subdomain/type/difficulty metadata and solution formatting. We examined ERI via seven LLMs and report a statistically significant three-tier performance structure, with frontier models (GPT-5, Claude Sonnet 4, DeepSeek V3.1) achieving mean scores above 4.30 on a five-point scale, while mid-tier and smaller models exhibited progressively higher failure rates and steeper performance degradation on graduate-level questions. To address circularity concerns inherent in LLM benchmarks, we developed a convergent validation protocol that leverages cross-provider independence, multi-judge averaging, and frontier-model agreement analysis to empirically bound hallucination risk to 1.7%. ERI is released with taxonomy specifications, validation scripts, and an evaluation harness to enable reproducible comparisons and regression testing for instruction tuning, routing, retrieval-augmented evaluation, and agentic tool-use workflows in engineering settings.

The dataset can be accessed from:
GitHub: https://github.com/mznaser-clemson/ERI-Benchmark
Hugging Face: https://huggingface.co/datasets/mznaser/ERI-Benchmark

## 1.0 Introduction

Engineering workflows increasingly rely on large language models (LLMs) and agents for various tasks (e.g., calculation support, design exploration, etc.). As such, LLM benchmarking has become a practical gatekeeper for model selection and deployment readiness [1]. While existing general benchmarks emphasize broad interpretation, engineering demands in-depth grounding and reasoning. For these reasons, an engineering-capable model must demonstrate not only procedural competence but also verification behavior, including the detection of underspecified inputs and constraint-aware reasoning that avoids silent violations.

One may then infer that a general-purpose benchmark is unlikely to be sufficient for engineering use. This insufficiency of general benchmarks becomes apparent when "correctness" is treated as more than arriving at a plausible answer. For example, the massive multitask language understanding (MMLU) benchmark measures broad multitask knowledge through standardized multiple-choice evaluation, yet its format does not support engineering tasks [2]. Similar benchmark suites, such as BIG-bench, probe diverse capabilities but neither enforce engineering-specific output contracts nor provide systematic intent-conditioned coverage across engineering subdomains [3]. Other, more specific benchmarks, e.g., math word-problem benchmarks (e.g., GSM8K), capture multi-step reasoning skills but are not structured around engineering fields and thus do not support automated engineering checks [4].

Thus, coverage in an engineering benchmark cannot be reduced to topic breadth alone. This is because engineering capability depends on the task's target, the required level of rigor, and the output format that downstream tools can reliably parse. So, a benchmark that pools heterogeneous prompts without explicit structure over engineering axes can produce strong average performance while obscuring systematic weaknesses in specific domains or task types. In such instances, evaluation results become aggregate scores that mask whether failures concentrate by rigor tier or by subdomain families that share aesthetic styles. Consequently, an engineering benchmark must frame coverage as a controlled cross-product of domains, task intents, and difficulty tiers [5,6].

From this lens, this paper introduces the Engineering Reasoning and Instruction (ERI) benchmark, a taxonomy-driven instruction dataset designed to train and evaluate engineering-capable LLMs and agents with controlled coverage across domains, task intents, and difficulty levels. ERI spans nine engineering fields and 55 subdomains, intersected with 7 intent types and 3 difficulty tiers. In total, ERI comprises 57,750 instruction–response records with explicit metadata. The selected taxonomy ensures engineering coverage, while intent types isolate the kinds of engineering work a model is asked to perform—ranging from definitions and explanations to calculations, comparisons, design guidance, troubleshooting, and code-oriented tasks. The adopted difficulty tiers encode how assumptions, edge cases, and verification expectations scale across tasks to support attribution of performance differences.

The remainder of this paper is organized as follows. Section 2 reviews related work on instruction tuning, domain benchmarks, and agent evaluation. Section 3 defines ERI's scope, intended users, and supported modeling workflows. Section 4 specifies the taxonomy, intent types, and difficulty rubric. Then, Section 5 details dataset construction, defines benchmark tasks, and the evaluation methodology. Section 6 reports baseline results from a multi-LLM study. Section 7 states limitations and responsible-use constraints, and Section 8 describes the release package and roadmap.

## 2.0 Background and related work

This section situates ERI within adjacent research threads that shape how engineering-capable LLMs are trained and evaluated. In addition, this section reviews instruction-tuning datasets and alignment methods, quantitative and scientific reasoning benchmarks, and agent/tool evaluations.

### *2.1 Instruction tuning datasets*

As noted above, heterogeneous instruction benchmarks can hide domain-specific weaknesses behind strong averages [7,8]. To overcome this limitation, instruction-tuning work moved evaluation beyond raw language modeling to reward adherence to directives, helpfulness, and dialogue-level stability [9]. Nevertheless, even under this framing, a model can learn to satisfy conversational expectations even when it fails to surface missing givens, fails to check dimensional consistency, or introduces untracked assumptions that would invalidate an engineering solution [10]. Hence, a benchmark that fails to capture these expectations tends to overestimate operational readiness.

One possible response to this gap is to scale coverage through synthetic instruction generation methods, which offer a path to broader coverage by automating prompt creation. Self-instructive style pipelines demonstrate that large instruction datasets can be expanded through model-generated tasks, thereby improving breadth and reducing annotation costs [11]. However, synthetic expansion typically inherits two structural limitations: 1) it often lacks a domain taxonomy that guarantees subdomain coverage, and 2) it rarely attaches formal output contracts that enable deterministic validation [12].

### *2.2 Domain benchmarks for quantitative and scientific reasoning*

Given that general instruction benchmarks lack engineering-specific structure, domain benchmarks seem like natural alternatives. For example, quantitative reasoning benchmarks in mathematics provide a valuable stress test for multi-step derivations and error-prone symbolic manipulation, and the math dataset remains a widely used example of this evaluation style [13]. However, mathematics benchmarks generally measure solution correctness in an abstract setting where physical constraints and/or feasibility checks are not central scoring objects. As one can see, even such benchmarks may not directly capture the verification behaviors that engineering workflows require.

This inference gap becomes more noticeable in scientific reasoning evaluations, given their emphasis on explanation quality and knowledge integration. Science-oriented benchmarks can probe conceptual coherence, which overlaps with parts of engineering reasoning in diagnosis and interpretation tasks. Even so, these benchmarks often accept free-form explanations as the primary output and rarely require structured constraint-satisfaction flags that can be automatically validated at scale. As a result, they provide signals about general reasoning competence but do not support intent-conditioned error analysis that would allow engineers to identify how failures arise.

This observation points to a narrower cluster of work that evaluates engineering-like performance using externally verifiable outcomes, most notably in software engineering (SWE). Such benchmarks evaluate models on real-world issues, with success defined as whether a generated patch resolves the issue at hand (thereby, offering a rigorous, operational target for code-capable systems) [14]. That rigor, however, is specialized to software tasks, and its scoring does not generalize to multi-field engineering, where a single pass/fail test suite does not often capture correctness. What emerges from this review is a need for a framework that borrows the operational discipline of task-based scoring while also encoding issues/tasks that dominate non-software engineering problems. Agent and tool-use evaluations represent one attempt to bridge this gap.

### *2.3 Agent and tool use evaluation*

The agent literature evaluates models as decision makers that alternate between reasoning and action [15]. For example, ReAct formalizes this alternation by coupling intermediate reasoning with explicit tool-oriented steps, which improves performance on tasks that require external interaction and multi-step control [8]. Despite this, the dominant agent evaluations often focus on success in navigation, retrieval, or API execution and primarily measure completion of outcomes. This creates a unique gap in which agents may complete tasks while still producing outputs that violate engineering constraints. This assessment gap reflects a broader methodological issue that cuts across benchmarks, since many evaluations conflate "produced an answer" with "produced a usable artifact."

### *2.4 The state of engineering-related datasets*

Recent engineering-facing LLM benchmarks continue to evolve, but the landscape remains fragmented across domains and scoring regimes. For example, AECBench frames capability along a cognition ladder (memorization→application), and anchors prompts in codes, tables, calculations, and professional document generation [16]. Another dataset, AECV-Bench, pushes drawing literacy by pairing floor plans with spatial reasoning [17]. BuildingQA extends grounding to BIM semantics, reframing "answering" as structured retrieval and relational reasoning over knowledge graphs [18]. In such datasets, scoring strategies range from exact-match to rubric-graded judging, which complicates fair cross-benchmark comparisons directly.

Building on this foundation, engineering course-centered benchmarks aim to isolate mechanistic brittleness that broader task mixtures can mask. In one dataset, SoM-1K demonstrates that even canonical strength-of-materials problems remain difficult when diagram interpretation is required, and that expert textual surrogates for visuals can outperform direct vision inputs, implying representation-dependent reliability rather than stable conceptual control [19]. The recent EngDesign dataset, which comprises about 101 tasks, validates design outputs through a simulation-based pipeline across multiple engineering domains [20].

Based on the above review, engineering settings demand outputs that can be checked, routed, and audited. Hence, evaluation requires both a structured response format and scoring components that detect silent failure modes. Without strict schemas, automated checking becomes fragile, and result comparisons become sensitive to superficial formatting differences or differences in rater interpretation. Further, today's benchmarks rarely normalize instruction intent (define vs explain vs compute vs troubleshoot) or difficulty in a way that supports reliable claims across engineering. ERI is positioned to close this gap by coupling an explicit instruction taxonomy with multi-field coverage and tiered difficulty.

## 3.0 Scope, intended users, and modeling use cases

In this section, we define what ERI is designed to measure, identify the audiences ERI is intended to serve, and outline the decisions the benchmark is meant to inform.

### *3.1 Scope boundaries and what ERI treats as measurable engineering competence*

ERI treats coverage as a controlled combination of field (civil, mechanical, electrical, chemical, environmental, aerospace, materials, fire, and industrial engineering), task intent (i.e., definition, explanation, calculation, comparison, design/synthesis, troubleshooting, and code-related), and difficulty tier (undergraduate, graduate, and professional). Each field–subdomain–intent–difficulty cell is populated with a fixed number of samples to prevent high-volume prompt classes from dominating overall scores and covering systematic weaknesses. In ERI, the benchmark generates 50 instruction–response pairs per cell, yielding 57,750 total records across 1,155 unique cells.

Such organization and comprehensiveness help conduct evaluations that can isolate whether failures concentrate in specific intents, rigor levels, or domain slices. ERI targets engineering tasks that can be expressed as instruction–response pairs and focuses on pre-decisional and midstream reasoning behaviors that appear across engineering fields (e.g., disciplined calculation, structured comparison, preliminary design rationale under stated constraints, and troubleshooting based on observable symptoms). That being said, ERI does not attempt to encode stamped design responsibility, enforce local code adoption choices, or certify compliance with any specific regulatory regime, because those tasks require context and liability structures that a benchmark cannot supply.

### *3.2 Intended users and the decisions ERI is meant to support*

ERI is designed for three overlapping user groups, each with a distinct decision surface. In particular:

- Researchers require a stable instrument for comparing models, diagnosing systematic errors, and running tests across model versions or prompting protocols. Thus, for researchers, the most actionable signal is often a slice-level structure that supports attribution, such as identifying that a model is strong on calculation intent but weak on troubleshooting intent at higher rigor tiers.
- Industry practitioners need evidence that a model can reliably produce constraint-aware outputs (given how deployment risk is often driven by silent failures rather than by visible incoherence). Herein, practitioners are usually interested in identifying failure rates to map to operational risk (e.g., contract noncompliance, unit inconsistency, or infeasible constraint handling under standardized prompts, etc.).
- Educators expect a way to support transparent critique of assumptions and verification behavior (since those behaviors shape how novices learn engineering reasoning norms). Thus, educators seek rubric dimensions that can be taught and assessed (e.g., whether the response states assumptions, checks results, and flags underspecification).

ERI is released with schema-validated records that follow a common instruction–response structure and carry explicit metadata, so users can run controlled evaluations, train competence routers, or perform targeted fine-tuning without rebuilding labeling infrastructure. At the same time, ERI's design assumes that different stakeholders will apply different acceptance thresholds and different cost constraints, which motivates the modeling use-case mapping that follows.

### *3.3 Modeling workflows supported by ERI*

In supervised fine-tuning or parameter-efficient adaptation workflows, ERI supports training that improves engineering-form output discipline within controlled domain slices, including settings where adapters or low-rank updates are preferred to full fine-tuning for cost and maintainability reasons [21]. Accordingly, ERI can serve as a curriculum-like resource for improving reliability in specific intent categories or difficulty tiers, while keeping the central claim grounded in evaluation: improvements are meaningful only to the extent that they reduce measurable engineering failure modes under the benchmark's scoring rules.

The aforenoted ERI's metadata enables competence profiling that supports routing decisions across field, intent, and difficulty. This becomes especially relevant when smaller LLMs/agents can handle lower-rigor slices at lower latency and cost, while larger models are reserved for high-rigor or high-ambiguity slices. This workflow is conceptually related to mixture-of-experts routing in large-scale modeling, where inputs are routed to different capacity pathways based on learned selection mechanisms [22]. In the same spirit, ERI can be used to evaluate agentic tool-use pipelines that combine language generation with calculators, solvers, or checkers, where the critical question is whether tool calls reduce engineering-relevant errors rather than merely improving stylistic confidence [23].

## 4.0 ERI design: taxonomy, task intents, and difficulty rubric

This section introduces the field–subdomain taxonomy that defines domain coverage, the intent types that specify the kinds of engineering work a model is expected to perform, and the difficulty rubric that determines how rigor and verification expectations scale.

### *4.1 Taxonomy for multi-field engineering coverage*

ERI's taxonomy has nine fields, each decomposed into subdomains that reflect distinct bodies of engineering knowledge and practice contexts. More specifically, ERI spans civil engineering, mechanical engineering, electrical engineering, chemical engineering, environmental engineering, materials engineering, aerospace engineering, industrial engineering, and fire science, with fifty-five subdomains distributed across those fields (see Table 1). It should be stressed that ERI does not argue that its subdomains exhaust each field's internal structure. Rather, the selected taxonomy is designed to supply stable anchors for problem generation and scoring aggregation as a means to track whether model updates improve competence uniformly or only in narrow slices.

Table 1 ERI's taxonomy

| Field | subdomains |
|---|---|
| Civil engineering | Statics, mechanics of materials, structural analysis, steel design, concrete design, geotechnical engineering, structural dynamics, construction management |
| Mechanical engineering | Thermodynamics, fluid mechanics, heat transfer, machine design, dynamics and vibrations, manufacturing, HVAC |
| Electrical engineering | Circuits, electronics, signals and systems, power systems, electromagnetics, control systems |
| Chemical engineering | Mass and energy balances, chemical thermodynamics, reaction engineering, separation processes, process control, transport phenomena |
| Environmental engineering | Water treatment, air quality, hydrology, waste management, environmental impact |
| Materials engineering | Structure of materials, mechanical properties, phase diagrams, failure analysis, polymers and composites, corrosion |
| Aerospace engineering | Aerodynamics, flight mechanics, propulsion, aerospace structures, orbital mechanics |
| Industrial engineering | Operations research, quality control, ergonomics, production planning, engineering statistics, reliability |
| Fire science | Fire dynamics, combustion fundamentals, structural fire engineering, wildland fire behavior, evacuation and human behavior, fire modeling and simulation |

*4.2 Task intent types and engineering-aligned output contracts*

ERI uses seven intent types that act as a second conditioning axis. This helps benchmark "knowing about" a domain against "doing" a domain (see Table 2) [24,25]. These evaluation criteria are engineered to distinguish between plausible narration and genuine reasoning. In particular, ERI rejects aesthetic or stylistic evaluation for DEF and EXP tasks—avoiding assessments of "good writing" in favor of rigorous verification that physical meaning is preserved and premises are not invented. Then, CALC, CMP, and DES operationalize the shift from conceptual understanding to disciplined execution under constraints by requiring models to manage procedures as opposed to retrieve facts.

Unlike the first five task types, TRB and CODE specifically probe failure modes that frequently lead to brittle breakdowns in deployed engineering systems (and hence require models to demonstrate not just domain knowledge but also metacognitive awareness of their own evidentiary limitations). The above architecture reflects how engineering work is actually reviewed in practice, in which the evaluator must be able to reconstruct the complete reasoning path and detect silent violations of constraints that might otherwise go unnoticed in generic text generation.

Table 2 ERI's task intent types

| Intent | Task Type | Requirement | ERI Success Criteria |
|---|---|---|---|
| DEF | Definition | Define concepts or core principles within a specific field–subdomain | Precision and explicit boundary conditions that prevent category errors |
| EXP | Explanation | Explain mechanisms linked to observed behavior | Causal or physical coherence; explicit clarification of assumptions when multiple regimes could apply |
| CALC | Calculation | Provide quantitative results | Present key steps for auditability while keeping derivation concise; report final answer with units when applicable |
| CMP | Comparison | Compare options or approaches | Name the comparison criteria explicitly; tie conclusions to those criteria rather than using preference language |
| DES | Design | Provide design-oriented guidance | Structured design approach including constraint recognition and defensible rationale for choices; avoid generic tip lists |
| TRB | Troubleshooting | Diagnose problems or failures | Organize hypotheses; propose discriminating checks between causes; avoid asserting single cause when symptoms are underdetermined |
| CODE | Code/Standard | Provide code or standard-oriented guidance | State relevant requirements in actionable, auditable way; acknowledge dependencies on jurisdiction, edition, or context when not provided |

*4.3 Difficulty tiers and the rubric for rigor and verification expectations*

Difficulty tiers define how demanding the work is expected to be within a given intent. Herein, ERI uses three tiers, labeled undergrad, graduate, and professional, and these tiers are supplied as explicit conditioning attributes during generation, where the dataset encodes the intended rigor level at the record level – see Table 3. For benchmarking, a model can be tested on a single tier or all tiers. This implies that a model's behavior can also be traced accordingly. So, a model that improves uniformly across tiers suggests improvements in core reasoning and verification habits.

In contrast, a model that improves only at lower tiers suggests better pattern matching without corresponding gains in rigor under constraint pressure.

Table 3 Difficulty tiers and the rubric within ERI

| Class | Tier | What the *item* should look like | Rubric checklist |
|---|---|---|---|
| DEF | Undergrad | Define a core term; basic context; maybe one governing equation | A, C; includes key properties/variables; avoids misconceptions; optionally 1 simple example |
| | Graduate | Definition + formalism (conditions/assumptions); connect to related concepts | A, S, C; includes scope/limits; gives formal definition + interpretation; contrasts near-neighbors |
| | Professional | Definition framed for practice (design implications, common misuse) | A, R, C; highlights practical consequences; “when it matters” + “when it doesn’t”; common field pitfalls |
| EXP | Undergrad | Explain "why/how" with one main mechanism | A, C; causal chain is coherent; uses correct terminology; avoids hand-waving |
| | Graduate | Mechanistic explanation with modeling lens (governing equations, parameter regimes) | A, M, S, C; identifies dominant mechanisms; notes regime changes/edge cases; states assumptions |
| | Professional | Explanation tied to diagnosis/decision (what to check, what it implies operationally) | A, R, C, T; links mechanism → observable signs → actions; includes constraints and practical checks |
| CALC | Undergrad | Plug-in to known formula or 2–4 step calculation | A, U, T, C; correct steps; correct units; final numeric answer; basic sanity check (V) encouraged |
| | Graduate | Multi-step derivation/model selection; possibly nondimensionalization/uncertainty | A, U, M, S, T, V; method justified; assumptions explicit; sensitivity/uncertainty at least mentioned |
| | Professional | Realistic calculation with constraints (availability of inputs, safety factors, code-style reporting) | A, U, R, T, V, C; inputs clearly stated; conservative choices justified; checks + recommendation-ready output |
| CMP | Undergrad | Compare two concepts/materials/methods on a few criteria | A, C; clear criteria; correct contrasts; simple recommendation if asked |
| | Graduate | Compare with deeper criteria (performance metrics, assumptions, failure modes) | A, M, S, C; notes tradeoffs; identifies where ranking flips under different assumptions |
| | Professional | Compare alternatives for a project decision (cost/schedule/risk/compliance) | A, R, C, T; decision matrix logic; constraints and risks explicit; final recommendation is defensible |
| DES | Undergrad | Design a simple component/system with stated targets; limited constraints | A, M, S, T, C; requirements restated; approach reasonable; basic sizing + justification |
| | Graduate | Design with modeling choices, multi-objectives, constraints; includes verification | A, M, S, T, V, C; shows rationale, constraints handling, and validation (check against limits) |
| | Professional | Design task mirrors practice: constraints, safety factors, constructability, compliance, tradeoffs | A, R, T, V, C; code/standard awareness where relevant; risks; implementation notes; clear final spec |
| TRB | Undergrad | Simple troubleshooting scenario (symptom → likely cause) | A, C; plausible cause(s); straightforward test/next step; avoids unsafe advice |
| | Graduate | Diagnosis plan: multiple hypotheses, discriminating tests, mechanism-based reasoning | A, M, S, T, C; structured hypothesis tree; tests prioritize information gain; notes confounders |
| | Professional | Field-ready triage: safety, immediate containment, verification, documentation | A, R, T, C; prioritizes safety; clear stepwise plan; "stop-work" thresholds if applicable |
| CODE | Undergrad | Small script/function; straightforward I/O; basic correctness | A, C, T; code runs conceptually; correct logic; minimal but clear comments; simple test/example |
| | Graduate | More robust code: validation, edge cases, numerical stability/performance awareness | A, M, T, V, C; tests; handles edge cases; explains algorithm choice and complexity |
| | Professional | Production-leaning snippet: reliability, logging, input checks, reproducibility, interfaces | A, R, T, V, C; defensive programming; clear API; reproducibility hooks; practical constraints (data, units) |

A = Accuracy/Correctness, U = Units/Dimensions, M = Method choice/justification, S = Stated assumptions/scope, V = Verification/sanity check, C = Clarity/structure, R = Realism/constraints (codes, safety, constructability), T = Traceability (inputs → steps → outputs)

*4.4 On the synthetic generation methodology and engineering-domain suitability*

ERI's adoption of an LLM-based synthetic generation methodology aligns with established practice in instruction-tuning research. In particular, the Self-Instruct framework demonstrated that LLM-generated instruction data can produce models with performance comparable to those trained on human-curated data, with Alpaca and subsequent datasets validating this approach at scale [11,26]. Further, recent surveys on synthetic data generation emphasize that domain-specific constraints and post-generation filtering can yield datasets that are "nearly as effective as real data," particularly when augmenting controlled evaluation settings [27]. ERI's taxonomy-constrained generation—where each record is seeded by an explicit (field, subdomain, intent, difficulty) tuple—imposes structural controls that distinguish it from unconstrained synthetic expansion and reduce the distributional drift that undermines purely self-instructed pipelines.

Engineering domains offer unique affordances that make synthetic generation more defensible than in open-ended linguistic tasks. First, engineering problems are governed by explicit correctness criteria and, often, a deterministic structure that must be satisfied. These properties enable partial automatic verification, a feature unavailable in creative/subjective domains. Second, engineering pedagogy has long relied on canonical problem formulations (e.g, standardized textbook problems that encode disciplinary norms for how constraints are stated, how solutions are structured, and how verification is performed [28]. Thus, a generative model trained on such data can naturally reproduce these norms. The recent EngTrace benchmark explicitly makes this argument, noting that its "entirely synthetic" dataset enables controlled evaluation of reasoning behavior without the confounds introduced by heterogeneous practitioner-authored problems [29].

## 5.0 Dataset construction, tasks, and evaluation

Here, we describe how ERI records are generated, stored, validated, filtered, and partitioned for downstream use.

*5.1 Record format, raw storage, and versioned release artifacts*

ERI's generator enumerates a seed key composed of the tuple (field, subdomain, intent type, difficulty) and uses it as the accounting unit for both coverage control and resumability. Each API call returns five instruction–response pairs for a given seed, and each seed is queried ten times to yield fifty total pairs per cell. To make these records machine-checkable at scale, the generator constrains model outputs to a strict JSON object whose only top-level key is "samples," where each element is an object with exactly two keys, "instruction" and "response." Once a batch is produced, each sample is written as a JSONL record that attaches the seed metadata fields required for analysis and evaluation. In particular, downstream validation expects each record to include "instruction," "response," "field," "subdomain," "type," and "difficulty" to enable self-describing records and avoid reliance on file naming conventions for semantics (see Fig. 1).

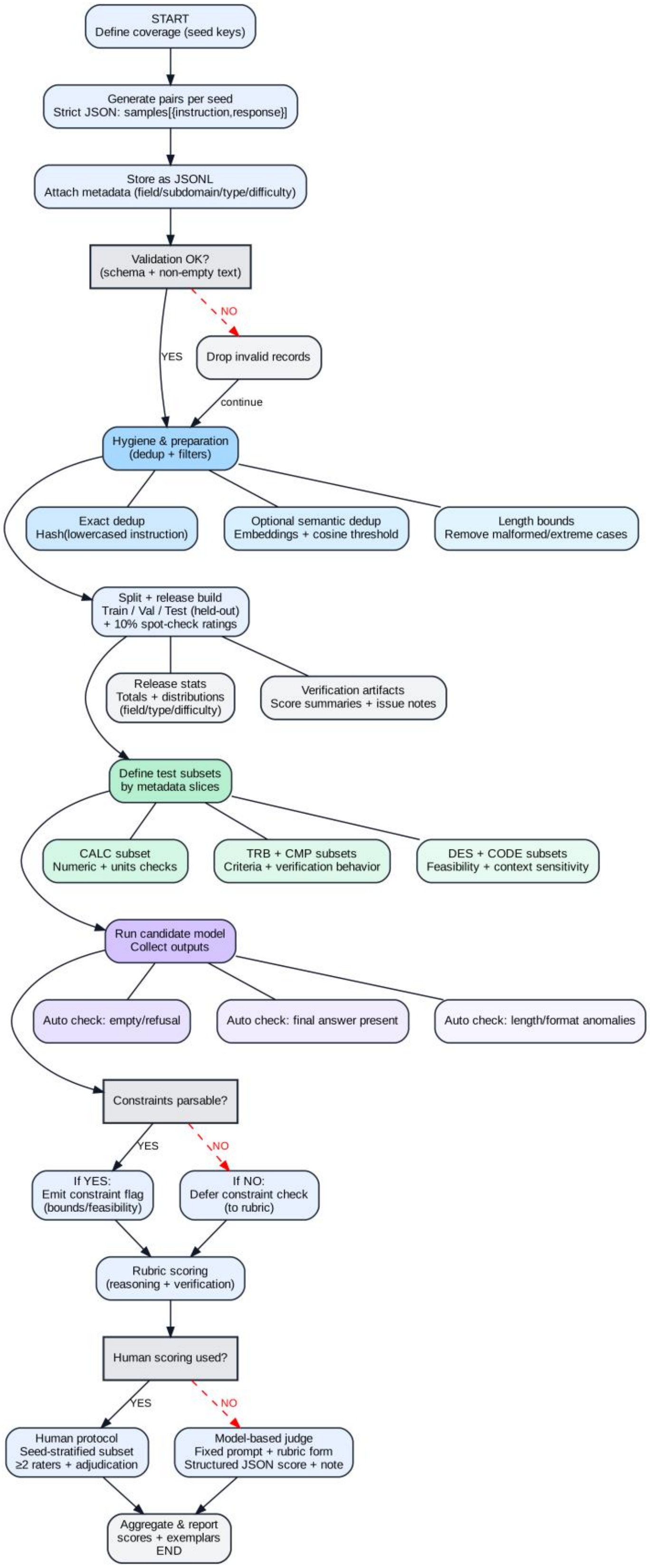

Fig. 1 Flowchart of ERI development

### *5.2 Post-processing: schema validation, deduplication, verification, and split construction*

The post-processing pipeline begins by consolidating all raw JSONL files into a single in-memory collection while explicitly excluding any error artifact files. Each record includes the required metadata keys and non-empty instruction and response strings; any record that fails these checks is placed in an invalid pool. After schema validation, the pipeline performs exact deduplication using a hash of the lowercased instruction text. This decision makes the instruction string the identity surface for duplicates, which matches the benchmark's goal of reducing prompt redundancy that can inflate apparent performance through repeated exposure to near-identical tasks.

Since exact deduplication cannot address paraphrased duplicates, ERI's pipeline also includes an optional semantic deduplication step that can be enabled when embedding tools are available. When activated, the script computes sentence embeddings for instructions and applies a cosine similarity threshold to identify and remove near-duplicates, using a Sentence-BERT-style encoder as the embedding backbone. Following deduplication, the pipeline applies length-based filtering to remove records that are likely to be malformed, trivially short, or unreasonably long for the intended instruction-tuning format.

Once structural and hygiene filters are applied, the ERI pipeline performs a targeted verification pass over a random subsample of the retained dataset. The pipeline selects a random 10% subset of records and asks an advanced LLM verification model (namely, GPT-5.1) to rate each instruction–response pair on a five-point quality scale, returning a JSONL object that includes either a numeric score and a brief issue description or null. This additional verification step is designed to provide an empirical check of content quality without requiring full human adjudication across tens of thousands of items and to produce summary statistics, such as mean and minimum scores, for spot-checking. Finally, the pipeline computes summary statistics reporting totals and distributions across fields, intent types, and difficulty tiers, along with average instruction and response lengths, and writes them to the release metadata.

### *5.3 Task subsets and standardized evaluation protocol*

ERI defines benchmark subsets by filtering the held-out split on the metadata axes introduced in Section 4, so each reported score can be fully traced to a field, subdomain, intent type, and difficulty tier. The default evaluation suite includes a full-coverage test pass that aggregates results across all slices, along with targeted diagnostic passes that isolate particular intents when a deployment or research question demands a narrower stress test. It should be noted that for calculation-only tasks, the evaluation suite includes a CALC-focused subset in which reference responses typically contain a numeric result and units to allow the scoring harness to attempt numeric extraction and unit consistency checks. For troubleshooting and comparison tasks, the suite includes TRB and CMP subsets where correctness depends more on hypothesis management, criteria articulation, and verification behavior. For design- and code-oriented tasks, the suite also includes DES and CODE subsets for feasibility and context-sensitivity tests.

### *5.4 Automatic checks*

The first layer of scoring applies automatic checks that operate directly on the model output text. As a baseline, the harness flags failures such as empty outputs, refusal patterns, or outputs that omit a final answer in intents that require one, and it also records length and formatting anomalies that indicate the model did not follow the expected response discipline. When the instruction includes explicit numeric bounds or feasibility conditions that can be parsed into simple predicates, the harness attempts to verify whether the model's proposed value or choice violates those bounds, and it emits a constraint-satisfaction flag when the check is applicable. When the constraint is not machine-parsable, the harness leaves this dimension to rubric scoring rather than forcing brittle heuristics that would distort comparisons.

### *5.5 Rubric scoring*

ERI also applies a rubric-based assessment layer that scores the model output on reasoning quality and verification behavior. The rubric is designed to measure engineering behaviors, including whether assumptions are stated and bounded, whether the response performs or recommends verification steps appropriate to the intent, whether constraints and feasibility conditions are respected, and whether the reaction avoids introducing unjustified premises that change the problem.

This structured judgment layer also requires a scalable rater protocol that remains stable across large test suites. ERI supports model-based judging in the style of established LLM-as-a-judge methodologies, where a strong judge model evaluates a candidate response against a rubric under a fixed prompt template and returns a structured JSON score plus a brief issue note [30]. ERI further adopts a form-filling approach to reduce free-form variance and to improve the traceability of rubric outcomes, which aligns with prior work that shows rubric-constrained, template-based LLM evaluation can correlate better with human judgments than unconstrained critiques [31]. This means that the rubric layer yields per-dimension scores that can be aggregated by field, intent, and difficulty.

When human scoring is used, ERI recommends a small, seed-stratified subset that spans multiple fields and intents, with two or more independent raters and a disagreement-resolution procedure that records the final adjudicated score and the reason for revision. When human scoring is not used, the judge prompt and settings remain fixed and versioned so that the conducted tests remain meaningful across model releases, and the benchmark reports judge-related artifacts, such as score distributions and failure exemplars, to support audit.

## 6.0 Benchmarking study

This section presents our methodology and results arising from benchmarking ERI.

### *6.1 Methodology*

We followed a three-stage evaluation protocol: 1) response generation, 2) multi-judge scoring, and 3) metric aggregation. Each of the seven benchmark models was evaluated on 10% of the full dataset, sampled stratified across engineering field, subdomain, intent type, and difficulty tier to preserve ERI's coverage while keeping inference and judging costs tractable. Subsampling at this scale is consistent with common practice in LLM evaluation, as noted in Self-Instruct [11], which created ~52K synthetic instructions but relied on a 252-item expert-written suite for human evaluation [1]. WizardLM similarly reports human evaluation on 218 real-world instructions [32]. Moreover, many widely used public benchmarks operate at comparable or smaller sizes, including MT-Bench (80 multi-turn questions) [33], GPQA Diamond (198 questions) [34], HumanEval (164 problems) [35], and AlpacaEval (805 prompts) [36]. Relative to these norms, ERI's 10% subset (that exceeds 5,000 items) provides substantially broader coverage while remaining economically feasible for multi-model, multi-judge benchmarking.

Scoring employed a panel of three LLM judges from distinct providers: Claude Haiku 4.5 (Anthropic), GPT-4.1 Mini (OpenAI), and Mistral Small 3 (Mistral). Each judge received the original question, the reference answer from the test set, and the model-generated response, then produced a structured score between 1 and 5, accompanied by a brief justification. The rubric defined five levels ranging from incorrect or irrelevant (1) through excellent with accurate and well-explained content (5). Judges returned responses in JSON format, which were parsed programmatically with fallback handling for malformed outputs. The final score for each model–item pair was computed as the arithmetic mean across all three judges (intended to attenuate individual judge biases while preserving sensitivity to genuine quality differences). The rationale for the selected multi-provider composition of the judge panel is that each judge model originates from a different organization with different training data and alignment procedures, and that systematic biases from any single provider are diluted in the aggregate score. In total, 115,962 judgments were made.

Overall, the inter-judge agreement was moderate to strong (see Fig. 2). In particular, Spearman correlations ranged from $\rho = 0.70$ (Claude Haiku 4.5 vs. Mistral Small 3) to $\rho = 0.80$ (Claude Haiku 4.5 vs. GPT-4.1 Mini), and within-one-point agreement rates exceeded 85% for all judge pairs. A separate judge-level analysis also revealed that Claude Haiku 4.5 scored 0.41 points below the grand mean of 3.84, functioning as the strictest evaluator, while Mistral Small 3 scored 0.29 points above it, serving as the most lenient. GPT-4.1 Mini exhibited near-neutral bias (+0.12). These offsets were consistent across all of the examined 7 models, which means that the relative ranking of models remained stable regardless of which individual judge was consulted. The three-judge averaging protocol adopted in this study was therefore effective at mitigating individual judge tendencies while preserving discriminative power between models.

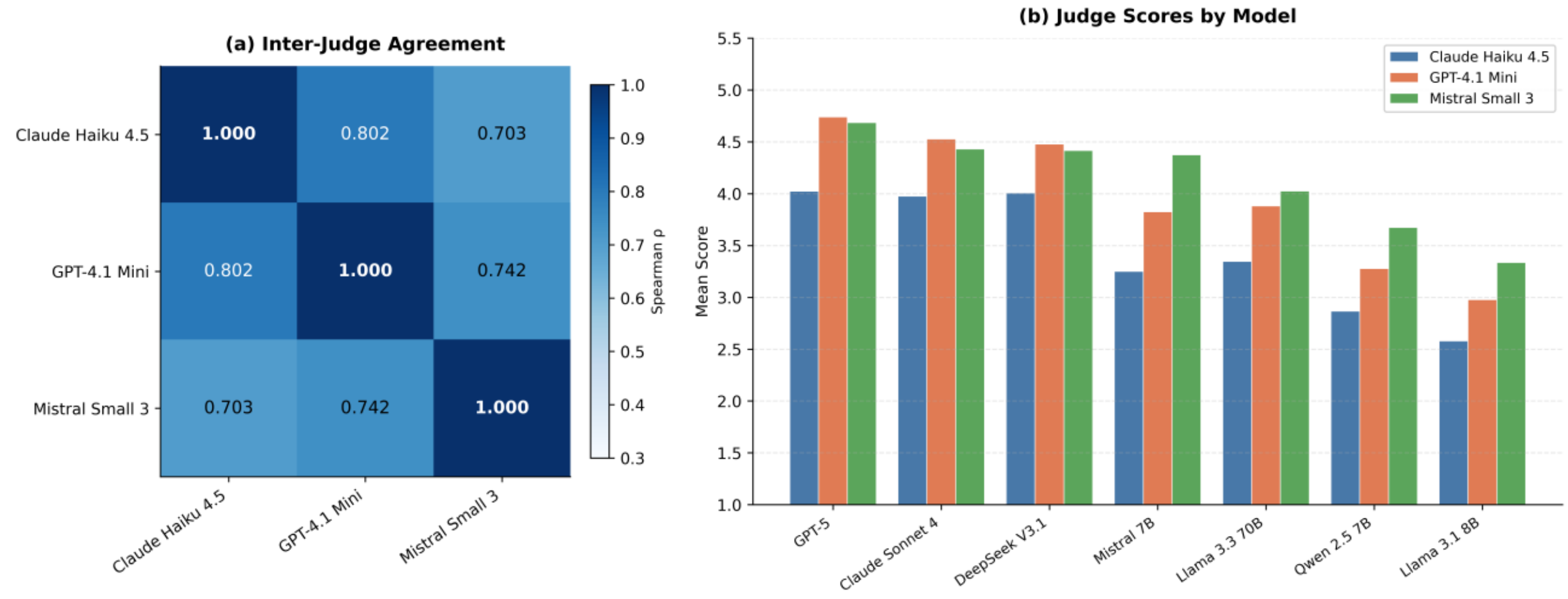

Fig. 2 Inter-judge agreement results

*6.2 Results*

Figure 3 shows the overall rankings for the tested LLMs. As one can see, GPT-5 achieved the highest mean score of 4.48 ($\sigma = 0.49$), followed by Claude Sonnet 4 at 4.31 ($\sigma = 0.66$) and DeepSeek V3.1 at 4.30 ($\sigma = 0.68$). In addition, a Kruskal–Wallis test was applied to confirm that differences among models were statistically significant ($p < 0.001$), and pairwise Mann–Whitney U tests with Bonferroni correction also identified significant separations between all pairs except one. The distinction between Claude Sonnet 4 and DeepSeek V3.1 was negligible in both practical and statistical terms (Cohen's $d = 0.02$, $p = 1.0$ after correction), which indicates that these two models perform at functionally equivalent levels on the ERI benchmark despite their substantially different architectures and training approaches. Bootstrap 95% confidence intervals further reinforce the precision of these estimates; the interval for GPT-5 spanned only 0.026 points (4.470–4.496), reflecting the statistical power afforded by over 5,000 scored items per model.

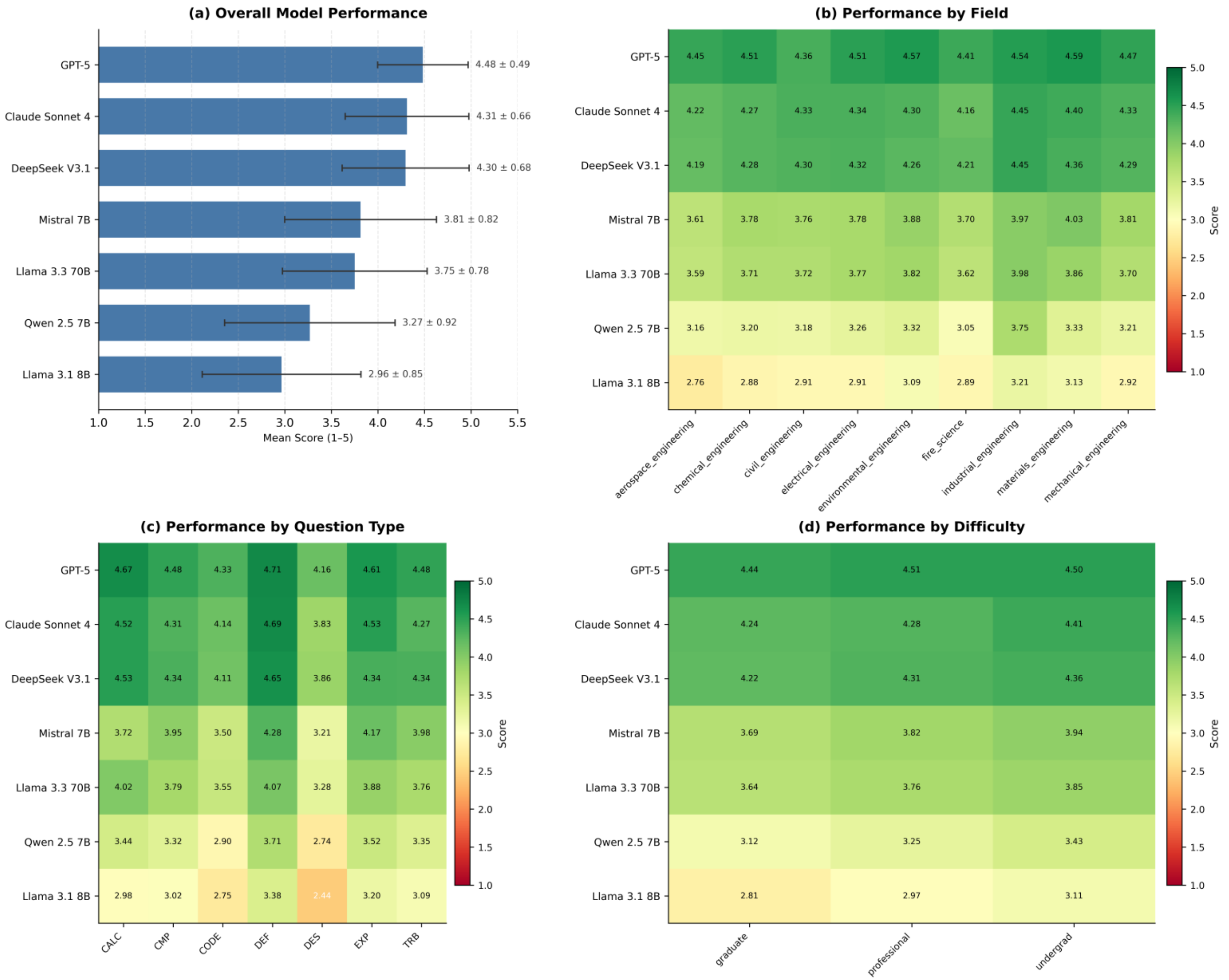

Fig. 3 Results of benchmarking

The mid-tier models comprised Mistral 7B (3.81) and Llama 3.3 70B (3.75), separated by a negligible effect size (d = 0.08). This result is interesting and warrants further attention because Llama 3.3 70B has 10 times as many parameters as Mistral 7B (which may suggest that parameter count alone does not determine performance on domain-specific engineering questions). The lower-tier models consisted of Qwen 2.5 7B (3.27) and Llama 3.1 8B (2.96), both of which exhibited failure rates (scores $\leq 2$) exceeding 10%, compared with less than 1% for frontier models. Score distributions further distinguished the aforementioned tiers: GPT-5 achieved a perfect score of 5 on 22.4% of items with only 0.1% failures, whereas Llama 3.1 8B reached a perfect score on just 3.5% of items while failing on 17.8% (see Fig. 4).

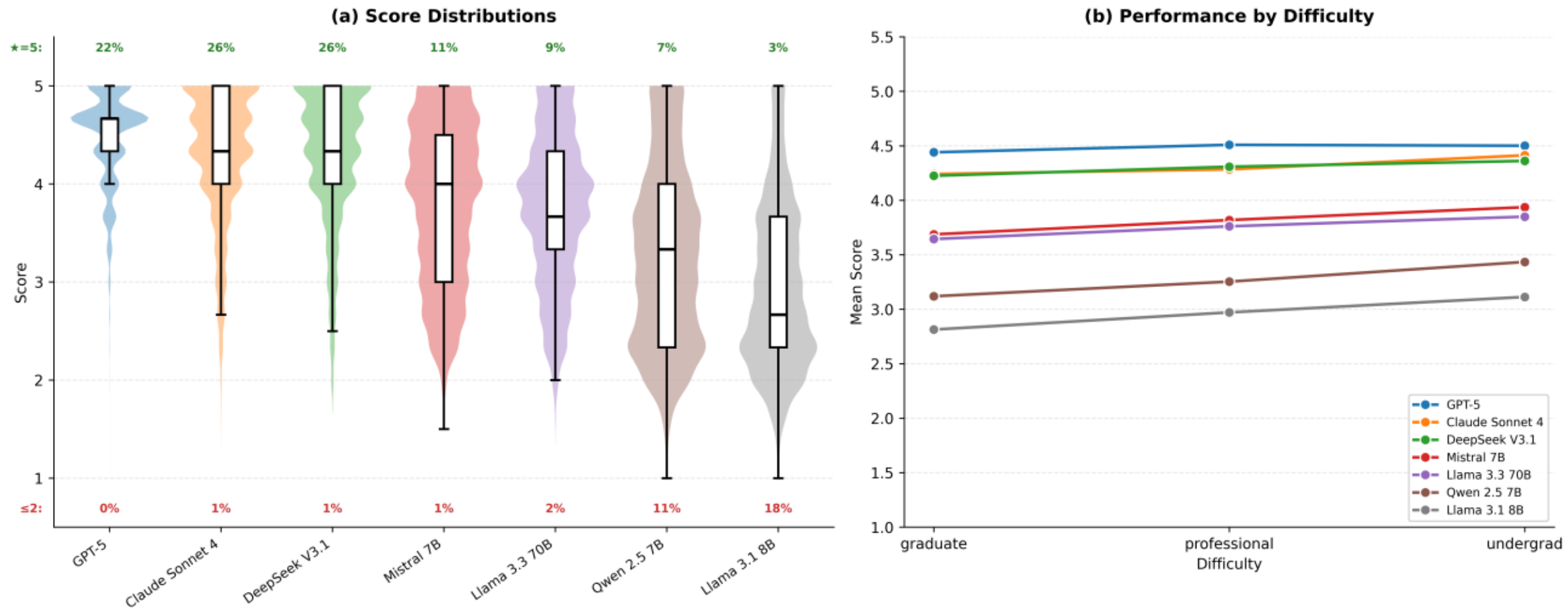

Fig. 4 Additional results of benchmarking

Performance varied meaningfully across question types. On the definition-based questions (DEF), these proved easiest for all models, with GPT-5 scoring 4.71 and even Llama 3.1 8B reaching 3.38. Design questions (DES) were consistently the most challenging and yielded the lowest scores across all models. The gap between the highest- and lowest-performing question types widened as model capability decreased. As shown in Fig. 4, the frontier models exhibited a spread of approximately 0.55 points between their best and worst question types, whereas smaller models showed spreads exceeding 0.93 points. Coding questions (CODE) similarly separated the tiers, with frontier models maintaining scores above 4.1 while the 7–8B models fell below 2.9, likely reflecting limited exposure to engineering-specific code during pretraining.

The cross-field analysis revealed that industrial engineering produced the highest scores across nearly all models, while aerospace engineering and fire science consistently ranked among the most difficult (see Fig. 3). The absolute spread across fields, however, remained modest. GPT-5 ranged from 4.36 (civil engineering) to 4.59 (environmental engineering), a span of 0.23 points, whereas Llama 3.1 8B ranged from 2.77 (aerospace) to 3.21 (industrial), a wider span of 0.44 points. This pattern suggests that weaker models are disproportionately affected by domain specificity, likely because their more limited parametric capacity provides uneven coverage across engineering subfields.

Figure 4 traces the difficulty across all models and shows a consistent, monotonic relationship, with scores decreasing from undergraduate to professional to graduate-level questions. The magnitude of this degradation, however, varied systematically with model capability. For example, GPT-5 lost only 0.06 points (1.4%) between undergraduate and graduate items, maintaining robust performance regardless of difficulty. By contrast, the 7–8B models lost 0.30–0.32 points (approximately 10%) – which may imply that smaller models disproportionately struggle as question complexity increases. This widening gap at higher difficulty levels carries practical implications for engineering education and professional applications, where graduate-level reasoning is often required.

### *6.3 The circularity argument*

A potential concern with any LLM-generated benchmark is the absence of human-verified ground truth. We address this through four complementary arguments:

- First, at the moment, the benchmark measures relative model performance rather than absolute correctness; systematic errors in reference answers affect all models equally and therefore do not distort rankings.
- Second, the generator model (GPT-5 Mini), the evaluated models (seven architectures from four providers), and the judge panel (three models from three providers) share no training pipeline. This breaks the circularity that would arise if a single system generated, answered, and evaluated its own questions.
- Third, we conducted a convergent validity analysis to empirically assess reference answer quality (see Fig. 5). Across all tested items, the three frontier models (i.e., GPT-5, Claude Sonnet 4, and DeepSeek V3.1) independently produced responses scoring 4.0 or above on 68.8% of items. This suggests that the reference answers for these items are factually sound. Had a reference answer been erroneous, at least one frontier model would have produced a response diverging from the gold standard, which judges would score lower. Further, cross-model standard deviations remained low overall (mean $\sigma = 0.74$, median $\sigma = 0.72$), and only 0.7% of items exhibited high disagreement ($\sigma > 1.5$) – which indicates that the benchmark contains few genuinely ambiguous or contested items. In fact, the 23.9% of items where at least one model scored 2.0 or below were concentrated among the 7–8B parameter models (which is consistent with capability limitations rather than reference answer deficiency).
- Fourth, to bound hallucination risk in the LLM-generated reference answers, we identified items where frontier models scored at or below 3.0—reasoning that high-capability models encountering a hallucinated reference would produce correct responses that judges penalize for disagreeing with the erroneous gold standard. Herein, only 1.7% of items met this criterion, and these flagged items were distributed roughly proportionally across all nine engineering fields (as opposed to being concentrated in a single domain). The score inversion rate (i.e., where smaller models outperformed frontier models, the strongest indicator that small models are parroting a hallucinated reference while frontier models correct it) was less than 0.1%. This rate provides an empirical upper bound on the extent of systematic hallucination contamination in the benchmark. Conversely, 41.8% of items received a mean score of 4.0 or above across all seven models, and 26.7% exhibited the expected capability gap pattern, in which frontier models scored above 4.0 while smaller models fell below 3.0, consistent with genuine difficulty differentiation rather than reference error.

Given the above, it is without doubt that future work will aim to supplement this analysis with expert validation of a stratified subsample against established engineering codes and standards to further root ERI with scaled human verification.

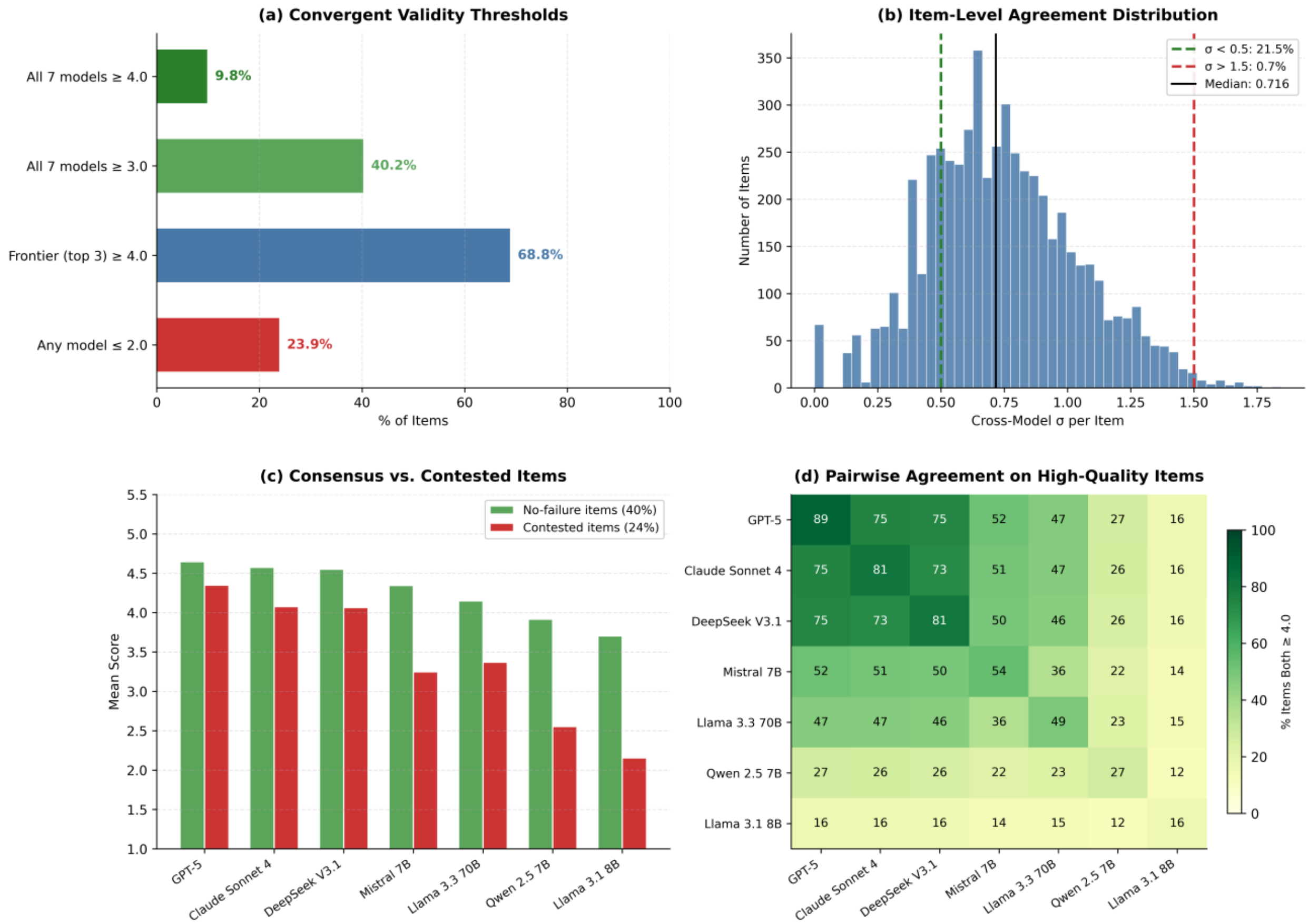

Fig. 5 Convergent validity analysis

## 7.0 Limitations and responsible use

This section sets out the limits of ERI's intended use and identifies risks arising from synthetic instruction data and metric-driven optimization, including bias, overfitting, and evaluation gaming.

### *7.1 What ERI cannot certify*

Engineering practice often requires jurisdiction-specific code interpretation, stamped responsibility, site conditions, and liability-aware judgment. Accordingly, ERI should not be used to claim compliance, safety certification, or readiness for autonomous control of high-stakes engineering decisions, and ERI scores should be interpreted as evidence about bounded behaviors defined by Section 4 and scored by Section 6 (vs. as a substitute for expert review). For this reason, ERI is best viewed as a controlled stress test that helps teams measure and reduce known failure modes, while leaving context acquisition, accountability, and final decision authority to human processes and domain governance frameworks.

### *7.2 Risks from prompting and metric optimization*

Relying on prompts can over-represent "clean" problem statements and, hence, under-represent messy real constraints. Thus, ERI should be treated as an evaluation and training resource that benefits from continuous auditing and augmentation, which aligns with broader dataset documentation norms that emphasize known gaps and intended-use limits rather than implicit claims of completeness. One should still realize that benchmark optimization can produce "teaching to the test" dynamics that inflate scores without improving robustness in the wild [39].

For that reason, ERI should be used with complementary checks such as out-of-distribution prompts, adversarially underspecified cases, and human review on critical slices, which fits risk-management guidance that treats evaluation as one control among several rather than as a sole assurance mechanism.

### *7.3 Temporal validity and standard evolution*

ERI's records reflect the knowledge state encoded in the generating model's training data at a fixed point in time, which means that code-related (CODE) and design-oriented (DES) items may reference superseded material. This temporal anchoring does not invalidate ERI for comparative benchmarking, where the goal is to measure relative model performance under controlled conditions, but it does limit ERI's utility as a reference for current professional practice. Organizations deploying ERI for fine-tuning or routing should audit slices that are sensitive to regulatory updates and plan periodic refresh cycles to regenerate or manually revise time-sensitive cells.

### *7.4 Evaluation coverage and silent omissions*

Despite spanning nine fields and fifty-five subdomains, ERI still does not cover other disciplines (e.g., biomedical engineering, nuclear engineering, software engineering beyond CODE-type items), and within included fields, coverage reflects subdomain salience at the time of taxonomy design rather than principled sampling of all professionally relevant topics. This boundary should be made explicit when reporting ERI results: a model that scores well on ERI's civil-engineering slices has demonstrated competence on the tasks ERI defines, not on the unbounded space of civil-engineering problems. Users extending ERI to new domains should follow the taxonomy specification and validation protocols described herein to maintain comparability and avoid introducing uncontrolled variation that would compromise cross-study reproducibility.

## 8.0 Release package and roadmap

This section specifies which artifacts are released with ERI and then describes the versioning and evaluation-harness interfaces that support controlled comparisons across models and over time.

### *8.1 Release artifacts for reproducible use*

ERI is distributed as JSONL splits, with each record self-describing through the metadata fields defined earlier. This way, users can filter and aggregate without relying on external file-naming conventions. This structure is paired with a taxonomy specification that serves as the authoritative reference for field and subdomain labels. The release includes schema validation scripts that check required keys and fundamental structural integrity, along with hygiene utilities that support exact deduplication and configurable semantic filtering when users extend the dataset. ERI also ships with an evaluation harness that reproduces the automatic checks and reporting structure described in Section 6. When either layer changes, ERI's reporting format is designed to make that change visible in metadata so longitudinal comparisons remain interpretable.

### *8.2 Roadmap for ERI-V2*

A few items come to mind for what would complement ERI. A first roadmap item is a human-verified "gold" subset that is small enough for expert review yet broad enough to cover multiple fields and intent types. Such a subset would provide a calibration anchor for rubric-based scoring without forcing full human labeling. A second item is a tool-trace extension that includes selected tool-use trajectories as references to enable the evaluation of agentic pipelines that claim reliability gains through structured action. A third item is a held-out subdomain protocol that tests transfer across unseen subdomains within a field, thereby strengthening claims about generalization beyond paraphrase robustness.

This roadmap also includes harder constraint regimes that increase the cost of hidden assumption drift. Examples include tasks with multiple interacting constraints, explicit feasibility regions, and adversarially underspecified prompts where safe behavior requires asking clarifying questions or refusing to over-specify. These additions would not replace ERI-V1's role as a baseline regression suite. Instead, they would create a layered benchmark family where V1 supports stable comparisons, and V2 supports sharper stress tests for deployment-oriented reliability.

## 9.0 Conclusions

The engineering profession lacks a comprehensive, multi-field benchmark for evaluating LLM performance on domain-specific reasoning tasks. Existing benchmarks either focus on general knowledge, target a single engineering discipline, or rely on multiple-choice formats that fail to capture the open-ended reasoning required in professional practice. The Engineering Reasoning and Intelligence (ERI) benchmark addresses this gap by spanning nine engineering fields, seven question types, and three difficulty levels across over 57,000 items. Thus, ERI provides the first large-scale evaluation instrument designed to measure LLM capability across the breadth of engineering knowledge and reasoning.

The ERI benchmark produced several findings with practical implications for engineering practice and education. A primary finding notes that frontier LLMs demonstrated near-expert performance across all nine engineering fields and seven question types, with GPT-5 losing only 1.4% of its score between undergraduate and graduate difficulty levels. This robustness suggests that current frontier LLMs can serve as reliable supplementary tools for engineering reasoning tasks across difficulty ranges. The negligible performance gap between Claude Sonnet 4 and DeepSeek V3.1 (Cohen's $d = 0.02$) further indicates that competitive performance on domain-specific engineering tasks is no longer confined to a single provider or architecture. At the same time, our findings also note steep degradation in much smaller LLMs (i.e., 7–8B-parameter models), with failure rates exceeding 10%. This suggests that smaller models remain unsuitable for safety-critical engineering applications without additional safeguards.

Beyond the benchmark itself, this study presents a replicable validation protocol for LLM evaluation datasets. The combination of cross-provider triangulation at each pipeline stage, convergent validity analysis through cross-model agreement, and hallucination bounding through frontier-inversion detection provides a methodological template that generalizes to other domains where human-curated ground truth is expensive or impractical to obtain at scale.

## Accessing the dataset

The ERI dataset is publicly available on HuggingFace and can be loaded directly in Python with no manual download required. Users first install the HuggingFace datasets library, then load any configuration with a single command:

```
# One-time installation
pip install datasets

# Load the full dataset
from datasets import load_dataset
ds = load_dataset("mznaser/ERI-Benchmark", "full")
```

```
# View the splits
print(ds)

# DatasetDict({
#    train: Dataset({features: ['instruction', 'response', 'field', 'subdomain', 'type', 'difficulty'], num_rows: 45415})
#    validation: Dataset({features: [...], num_rows: 2671})
#    test: Dataset({features: [...], num_rows: 5343})
# })

# Access a single sample
sample = ds["train"][0]
print(sample["instruction"])          # The engineering question
print(sample["response"])             # The reference answer
print(sample["field"])                # e.g., "civil_engineering"
print(sample["type"])                 # e.g., "CALC"
print(sample["difficulty"])           # e.g., "graduate"

# Load a specific configuration
ds = load_dataset("mznaser /ERI-Benchmark", "civil")              # Civil engineering only
ds = load_dataset("mznaser /ERI-Benchmark", "calculations")   # CALC type only
ds = load_dataset("mznaser /ERI-Benchmark", "graduate")          # Graduate difficulty only
ds = load_dataset("mznaser /ERI-Benchmark", "small")              # 10K random subset

# Convert to pandas for analysis
df = ds["train"].to_pandas()
```

GitHub: https://github.com/mznaser-clemson/ERI-Benchmark

Hugging Face: https://huggingface.co/datasets/mznaser/ERI-Benchmark

## Data availability

Data is available on request from the author.

## Conflict of interest

The author declares no conflict of interest.

## Funding declaration

None.